\documentclass{article}
\usepackage{kr}

\usepackage{times}
\usepackage{soul}
\usepackage{url}
\usepackage[hidelinks]{hyperref}
\usepackage[utf8]{inputenc}
\usepackage[small]{caption}
\usepackage{graphicx}
\usepackage{amsmath}
\usepackage{amsthm}
\usepackage{booktabs}
\usepackage{algorithm}
\usepackage{algorithmic}
\usepackage{comment}
\usepackage[disable,colorinlistoftodos]{todonotes} 
\usepackage{natbib}
\usepackage{fontawesome}
\usepackage{flafter}

\newtheorem{example}{Example}

\title{FC-CONAN: An Exhaustively Paired Dataset for Robust Evaluation of Retrieval Systems}

\author{%
Juan Junqueras$^1$\thanks{Corresponding author: \texttt{jjunqueras@dc.uba.ar}}\and
Florian Boudin$^2$\and
May Myo Zin$^3$\and
Nguyen Ha Thanh$^{3,4}$\and
Wachara Fungwacharakorn$^3$\and
Damián Ariel Furman$^1$\and
Akiko Aizawa$^4$\and
Ken Satoh$^{3,4}$ \\
\affiliations
$^1$Universidad de Buenos Aires, FCEyN, Departamento de Computación, Buenos Aires, Argentina\\
$^2$JFLI, CNRS, Nantes University, Nantes, France\\
$^3$Center for Juris-Informatics, ROIS-DS, Tokyo, Japan\\
$^4$National Institute of Informatics (NII), Tokyo, Japan
}
\begin{document}

\maketitle

\begin{abstract}
Hate speech (HS) is a critical issue in online discourse, and one promising strategy to counter it is through the use of counter-narratives (CNs). Datasets linking HS with CNs are essential for advancing counterspeech research. However, even flagship resources like CONAN~\citep{chung-etal-2019-conan} annotate only a sparse subset of all possible HS–CN pairs, limiting evaluation. We introduce \textbf{FC-CONAN} (Fully Connected CONAN), the first dataset created by exhaustively considering all combinations of 45 English HS messages and 129 CNs. A two-stage annotation process involving nine annotators and four validators produces four partitions—Diamond, Gold, Silver, and Bronze—that balance reliability and scale. None of the labeled pairs overlap with CONAN, uncovering hundreds of previously unlabelled positives. FC-CONAN enables more faithful evaluation of counterspeech retrieval systems and facilitates detailed error analysis. The dataset is publicly available\footnote{The dataset is publicly available at \url{https://github.com/jnqeras/FC-CONAN-dataset}}\footnote{This work was partially completed while the first author was at the National Institute of Informatics (NII), Tokyo, Japan.}.

\end{abstract}

\section{Introduction}
\label{sec:Introduction}
\todo{Must: chequear todas las citas.}
\todo{Must: buscar mejor titulo.}
\todo{Must: agregar a la intro que pongo disponibles todos los datasets que usamos para los experimentos (por ahora solo voy a poner el experimento de recomendacion).}
\todo{Must: en la version para el workshop no voy a incluir la seccion sec:textGenerationModels, por lo que voy a tener que eliminar toda referencia a esta seccion y buscar toda referencia a text generation del paper.}
\todo{must: el dataset de Gurevych de dataset quality lo tengo aca y se llama 'Analyzing Dataset Annotation Quality Management in the Wild': https://drive.google.com/drive/folders/1ya2AbnZsDo7Ne-d1FhBN7ZovWxEthI0e}
\todo{This todo is not related with this section, but is to mention that I've done a thorugh researc in finding papers that do experiments of information retrieval and/or text generation using the CONAN dataset and I've found none (I even asked the author of the CONAN dataset) so it's safe to say that there is no such experiment, therefore, we will have to do experiments by utilizing SOTA methodologies for text generation and reccomendation or information retrieval utilizing the CONAN dataset and our new dataset and then compare the results.}
\todo{Must: Sacar referencias para citas del paper original del CONAN}
\todo{Must: Este paper puede tener buenas referencias para incluír: Directions for NLP Practices Applied to Online Hate Speech Detection}
\todo{Must: Agregar referencias a los papers del grupo de Argumentacion.}
\todo{Must: Check in the entire paper that I've written hatespeech and counter-narrative the same way.}
\todo{Must: For reproducibility, it is suggested to not only publish the adjudicated corpus but also raw annotations by the respective annotators. - sacado del paper de Gurevich.}

\noindent\textbf{Disclaimer.} This paper quotes hate speech verbatim for research purposes; some readers may find the language offensive.

Many Natural Language Processing (NLP) datasets consist of paired sentences, such as questions and answers \citep{rajpurkar-etal-2016-squad}, paraphrases \citep{dolan2005automatically}, entailment \citep{bowman-etal-2015-large}, translation \citep{koehn-2005-europarl}, and dialog \citep{li-etal-2017-dailydialog}. While some datasets allow a single sentence to link with multiple others, such as CONAN~\citep{chung-etal-2019-conan}, most cover only a fraction of all possible combinations. Exhaustive annotation is rarely attempted due to combinatorial growth and cost, so unlabeled pairs remain ambiguous—often reflecting oversight rather than a true absence of relation. This incompleteness is especially problematic for recommendation tasks, where metrics can severely underestimate system performance (§\ref{sec:ExperimentalEvaluationEvaluationOfRecommendationSystems}).

This challenge becomes particularly acute in domains such as hate speech. Social media has amplified the spread of harmful rhetoric \citep{Silva_Mondal_Correa_Benevenuto_Weber_2021}, \citep{waseem-hovy-2016-hateful}, prompting responses beyond content removal, which can reinforce censorship narratives. As a more constructive alternative, structured counterspeech protocols focus on timely, thoughtful responses that dismantle harmful arguments, avoid fostering further conversations, and align with broader goals. The CONAN dataset (COunter NArratives through Nichesourcing)~\citep{chung-etal-2019-conan}, "the first large-scale, multilingual, expert-based dataset of hate speech/counter-narrative pairs", remains the primary resource. To illustrate the nature of this data, we provide a representative example below.

\begin{example}[An HS–CN pair from CONAN~\citep{chung-etal-2019-conan}]
\textbf{HS (hate speech)} ``\textit{I hate Muslims. They should not exist.}'' \\
\textbf{CN (counter-narrative).} ``\textit{Muslims are human too. People can choose their own religion.}’’\\
\end{example}
Despite its widespread use, CONAN has notable limitations. \textbf{Crucially}, it does \emph{not} annotate all HS–CN combinations, leaving many appropriate pairs unlabeled. This limits its usefulness, specifically for evaluating CN recommendation systems, as performance metrics reflect only a lower bound. In a pilot study using one of these systems, we found that while only 2 of 10 suggested CNs were labeled as appropriate, manual review judged 8 to be valid—highlighting the risk of underestimating system accuracy.

The lack of full HS–CN pair annotations also limits generation tasks by reducing training data for fine-tuning LLMs. Comprehensive HS–CN annotations would further enable methods such as contrastive learning. Ultimately, unannotated pairs leave valuable latent information unused, decreasing the dataset’s utility for downstream applications.

Another characteristic of the dataset is that the guidelines are rather open-ended. This approach stems from the fact that the original annotators had already been trained to follow NGO guidelines for crafting effective CNs. These guidelines are notably consistent across both languages and organizations, and closely mirror those established in the Get the Trolls Out project\footnote{\url{https://getthetrollsout.org/stoppinghate}}. Annotators were encouraged to rely on their intuition, avoid overthinking, and produce reasonable responses \citep[§3.2]{chung-etal-2019-conan}. It's important to note that the high level of subjectivity is a characteristic of this field.

Due to resource constraints, we re-annotated a representative \emph{subset} of all possible HS–CN pairs. This paper details that effort, originally motivated by the need to evaluate a counterspeech recommender more accurately. While CONAN covers three languages, our work focuses solely on English. Extending the annotation to other languages is future work.

\section{Related work}
\label{sec:RelatedWork}
\todo{Puedo agregar quote a mis notas, sobre todo a la conclusion, del paper que tengo en mi drive: CrowdTruth 2.0: Quality Metrics forCrowdsourcing with Disagreement. Sobre esto, tener en cuenta mi reflexion que le envie a satoh sensei en un mail: he metrics proposed in the paper are primarily designed for crowdsourcing, while our study employed nichesourcing. Nevertheless, it would still be possible to calculate some of the metrics suggested. However, a major limitation is that many of these metrics assume — or are most meaningful — when all or a substantial number of annotators share a common set of examples. In our case, each example was annotated by at most three annotators, and the majority were annotated by one.

There is one calculation that could be adapted to measure agreement between two annotators. However, under these circumstances, it might be more appropriate to simply compute the inter-annotator agreement directly.

For these reasons, I believe that using the metrics from this paper would not offer a significant advantage. Nonetheless, I might still be able to reference it in the discussion of our paper. }

In the hate speech domain, CONAN \citep{chung-etal-2019-conan} is among the best-known multilingual resources. Expert-curated and focused on Islamophobia, it features hate speech (HS) and counter-narrative (CN) pairs in English, Italian, and French. Initially, included 4,078 pairs (1,288 in English) based on 136 unique HS messages, each matched with an average of 9.5 CNs. Through translation and paraphrasing, the English portion was expanded to 6,654 pairs, with 408 unique HS messages and 1,270 CNs. The dataset also includes metadata such as expert demographics, CN type, and HS sub-topic.

Several other datasets focus on hate speech and counter-narratives, such as DIALOCONAN \citep{bonaldi-etal-2022-human}, which features multi-turn dialogues between a hater and an NGO operator, though it is not organized in HS/CN pairs. Another example is Multitarget CONAN \citep{fanton-2021-human}, a dataset with HS/CN pairs addressing multiple targets of hate. However, these datasets do not consider every possible combination of HS and CN pairs. From \citep{furman-etal-2023-high}, one finding is most relevant to our work: a small LLM fine-tuned on a few hundred high-quality HS–CN pairs can outperform larger models. Our work complements theirs by focusing not on argumentative cues, but on exhaustively pairing HS–CN examples for robust evaluation and training.

A number of datasets have been developed for the task of hate speech detection, such as the Twitter corpus introduced by \citep{waseem-hovy-2016-hateful}. Although our dataset could potentially be utilized for this purpose, it is primarily designed with a different focus.

\section{Dataset Creation}
\label{sec:DatasetCreation}

We present FC-CONAN, a dataset of HS–CN pairs derived from a subset of the CONAN corpus. During annotation, we exhaustively considered all possible combinations of selected HS and CN items, labeling each pair as \emph{appropriate} or \emph{non-appropriate}, with some removed based on quality control criteria.

From the English partition of the CONAN dataset (the only language common to all annotators), we randomly selected 45 HS messages and collected the 375 CNs originally paired with them. For each CN, we used the SBERT model \texttt{all-MiniLM-L6-v2}\footnote{\url{https://huggingface.co/sentence-transformers/all-MiniLM-L6-v2}} to retrieve its two most similar CNs from the full CONAN dataset. We then discarded the 375 original CNs while retaining the HS messages and their newly retrieved CNs, ensuring that only novel HS–CN combinations were kept, so that annotators worked with CNs similar to the originals but not identical, preventing the task from being too easy. This process resulted in 133 unique CNs, from which we randomly selected 129 to ensure an even distribution of HS–CN pairs across annotators, producing 5,805 HS–CN pairs. Nine annotators—academically trained volunteers—labeled these pairs following adapted CONAN guidelines: fact-based information and maintaining a non-offensive tone to avoid escalating the conversation. Labels were assigned independently, with overlap enabling inter-annotator agreement checks, and adjudication resolved conflicts. Pairs marked \emph{not sure} or irreconcilably disputed were discarded to avoid label bias, leaving 5,032 labeled pairs (4,143 as \emph{‘the CN is not appropriate for the given HS’} and 889 as \emph{‘the CN is appropriate for the given HS’}). 

To further ensure reliability, 4,000 adjudicated pairs underwent a validation round by four independent reviewers (not involved in initial annotation), each re-assessing 1,000 pairs. Validators applied the same guidelines, could skip distressing items (three pairs were skipped), and judged whether labels conformed. This process ensured every retained pair has both annotator and validator input. The validated pairs were retained, regardless of whether they were deemed valid or not. Pairs outside this set were discarded to prioritize label reliability over dataset size. Although some excluded pairs may still be appropriate, all possible combinations were reviewed during annotation.

The final resource balances reliability and coverage by defining four quality-graded subsets (\emph{Diamond}, \emph{Gold}, \emph{Silver}, \emph{Bronze}), allowing users to trade size for label confidence. Ethical safeguards included warnings, optional skipping, and on-demand debriefing breaks to support annotators. The dataset may pose dual-use risks if inverted to generate hateful replies; we therefore stress the need for responsible downstream use.

\section{Results / Analysis.}
\label{sec:Results}
We begin by describing the dataset itself before transitioning to system-level evaluation. Section \ref{sec:DatastPartitions} introduces the four quality-graded partitions generated through our annotation and validation pipeline. Section \ref{sec:QualitativeAnalysisOfTheDataset} then explores their internal structure. Finally, Section \ref{sec:ExperimentalEvaluationEvaluationOfRecommendationSystems} presents a retrieval experiment that quantifies the impact of these partitions on downstream performance.

\subsection{Dataset Partitions.}
\label{sec:DatastPartitions}

\begin{table}[t]
\centering
\begin{tabular}{lrrr}
\toprule
Partition & Total & Appr. & Non-Appr. \\
\midrule
Diamond & 551  &  35 &  516 \\
Gold    & 663  &  54 &  609 \\
Silver  & 3580 & 431 & 3149 \\
Bronze  & 3997 & 702 & 3295 \\
\bottomrule
\end{tabular}
\caption{Number of hate speech -- counter-narrative (HS--CN) pairs in each dataset partition, categorized by appropriateness.}
\label{tab:partitions-appropriate-non-appropriate}
\end{table}

Following the annotation and validation  processes (§\ref{sec:DatasetCreation}), we obtained HS–CN pairs annotated by one to three annotators and assessed for validity by one validator. Based on whether annotator labels aligned and the results of the validation process, we defined four distinct dataset partitions, each differing in annotation quality and size. Table~\ref{tab:partitions-appropriate-non-appropriate} summarizes the size of each partition and the distribution of appropriate vs. non-appropriate HS–CN pairs. The characteristics of each partition are detailed below:

\begin{itemize}
    \item \textbf{Diamond Standard Dataset:} This partition includes only HS-CN pairs annotated by two or more annotators who reached unanimous agreement --whether the counter-narrative was deemed appropriate or non-appropriate. Additionally, a validator has confirmed the accuracy of these annotations.

    \item \textbf{Gold Standard Dataset:} This partition extends the \textit{Diamond Standard Dataset} by incorporating additional HS-CN pairs annotated by two or more annotators, regardless of whether the annotators unanimously agreed. In cases where annotators initially disagreed, these disagreements were resolved through the adjudication process (mentioned in §~\ref{sec:DatasetCreation}). Each resulting annotation was further reviewed and confirmed as accurate by a validator.

    \item \textbf{Silver Standard Dataset:} The \textit{Silver Standard Dataset} includes all HS-CN pairs from the \textit{Gold Standard Dataset}, along with pairs annotated by only one annotator and subsequently confirmed by the validators. Thus, annotations in this partition come from 1 to 3 annotators, with initial disagreements resolved via the aforementioned adjudication process. All annotations in this partition were approved during the validation stage.

    \item \textbf{Bronze Standard Dataset:} This dataset comprises all entries from the \textit{Silver Standard Dataset}, supplemented by all the HS-CN pairs that were not approved during the validation stage. As before, annotations originate from 1 to 3 annotators, with disagreements resolved through the adjudication phase. However, unlike the Silver Standard, this partition also retains pairs that were not approved during the validation stage.
\end{itemize}

As expected, these partitions differ inversely in size and quality. Higher-quality datasets (Diamond and Gold) require greater annotation agreement and validation, resulting in smaller dataset sizes. Conversely, lower-quality datasets (Silver and Bronze) are larger but contain annotations with potentially reduced reliability. Thus, the datasets, arranged from smallest to largest (highest to lowest quality), are: Diamond, Gold, Silver, and Bronze Standard datasets.

\subsection{Qualitative Analysis of the Dataset}
\label{sec:QualitativeAnalysisOfTheDataset}
\todo{Must: nicetohave: gregar comparacion del kohen Kappa contra el CONAN (el conan dice que tiene un 55 porciento o algo asi).}
\todo{Nice to have: (lo dejo en nice to have porque no puedo agregar mas cosas para el workshop) agregar los plots de las funciones plot\_bar\_counts y plot\_pie\_counts de la notebook esta seccion}
\todo{Must: tener en cuenta para el inner annotator agreement: That said, you could consider reporting the range or standard deviation along with the average to give a better sense of variability. If the number of groups were larger, Fleiss' Kappa or Krippendorff's Alpha would be alternatives that allow for multi-rater agreement. But in your case (groups of two and one of three), Cohen's Kappa remains appropriate.}
\todo{Esto ya lo agregue: en la notebook calculo el cohen kappa con la función, compute_cohen_kappa_for_all_annotator_pairs, tegno que reportarlo. 
Este es el resultado de esa funcion antes de que intente hacer cualquier cambio con respecto a los pares anotados con 1, creo que los cambios que haga alos pares anotados con 1 no deberian afectar en nada porque esos cambios se hacen dps de calcular el kohen Kappa:

Computing Cohen's kappa score for annotators Juan-san and Ken-san...
Annotator A: Juan-san, Annotator B: Ken-san, Cohen's kappa score: 0.3807339449541285
Computing Cohen's kappa score for annotators Florian-sensei and Leane-san...
Annotator A: Florian-sensei, Annotator B: Leane-san, Cohen's kappa score: 0.5340949660835416
Computing Cohen's kappa score for annotators Jonas-san and Leane-san...
Annotator A: Jonas-san, Annotator B: Leane-san, Cohen's kappa score: 0.5163527100427598
Computing Cohen's kappa score for annotators Florian-sensei and Jonas-san...
Annotator A: Florian-sensei, Annotator B: Jonas-san, Cohen's kappa score: 0.4149683766690091
Computing Cohen's kappa score for annotators Tom-san and Jiahao-san...
Annotator A: Tom-san, Annotator B: Jiahao-san, Cohen's kappa score: 0.05061410459587956
Computing Cohen's kappa score for annotators Julian-san and Xanh-san...
Annotator A: Julian-san, Annotator B: Xanh-san, Cohen's kappa score: 0.15421686746987961
[('Juan-san', 'Ken-san', np.float64(0.3807339449541285)),
 ('Florian-sensei', 'Leane-san', np.float64(0.5340949660835416)),
 ('Jonas-san', 'Leane-san', np.float64(0.5163527100427598)),
 ('Florian-sensei', 'Jonas-san', np.float64(0.4149683766690091)),
 ('Tom-san', 'Jiahao-san', np.float64(0.05061410459587956)),
 ('Julian-san', 'Xanh-san', np.float64(0.15421686746987961))]
}
\todo{Nice to have: (lo dejo en nice to have porque no puedo agregar mas cosas para el workshop): puedo agregar los analisis y plots que estan en la notebook a partir de "df\_sum\_total\_of\_annotations", la gracia de esto es que es un analisis de cada uno de los anotadores.}
\todo{Must: en esta seccion solo quedan "nice to have", los que realmente importan son los que tienen la marca "really nice to have".}

Across the six overlapping subsets described in Section~\ref{sec:DatasetCreation}, we obtain a mean Cohen’s $\kappa = 0.34$ ($\sigma \approx 0.20$), computed \emph{before} any additional checks. As noted by Klie et al., ``\emph{although it is often treated as such, agreement is no panacea; high agreement does not automatically guarantee high-quality labels.}’’ \citep{klie-etal-2024-analyzing}, so we applied the validation procedure outlined in Section~\ref{sec:DatasetCreation}. This section presents a qualitative analysis of the four dataset partitions: Diamond, Gold, Silver, and Bronze. Understanding these partitions helps users select the most suitable subset—prioritizing annotation reliability (Diamond/Gold) or volume (Silver/Bronze). \todo{must: en realidad es un nice to have, me gustaria tener una cita que diga que hate speech es highly subjective y que diga que Cohen's k esta ok para este tipo de tarea.}
\todo{Nice to have: comento esta imagen porque me di cuenta que la suma de las cantidades de cada una de las columnas de este plot no da el total de pares compartidos por los anotadores.}

\todo{Los nice to have mas importantes de esta seccion, los marque con "Muy importante nice to have"}
    
\begin{table}[t]
\centering
\begin{tabular}{lrr}
\toprule
Partition & Valid & Non-Valid \\
\midrule
Diamond &  551 &   0 \\
Gold    &  663 &   0 \\
Silver  & 3580 &   0 \\
Bronze  & 3580 & 417 \\
\bottomrule
\end{tabular}
\caption{Number of hate speech -- counter-narrative (HS--CN) pairs in each dataset partition, categorized by validity.}
\label{tab:partitions-valid-non-valid}
\end{table}

Table~\ref{tab:partitions-appropriate-non-appropriate} confirms that many valid HS–CN pairs were unannotated in the original CONAN dataset. By evaluating new combinations, we recovered hundreds of appropriate and inappropriate pairs across all partitions.
\todo{Muy importante nice to have: Hacer un count distinct de HS y otro de CN en cada una de las particiones Diamond, Silver, Gold y Bronze.}

Table~\ref{tab:partitions-valid-non-valid} shows a trade-off between quality and quantity: Diamond and Gold are smaller but fully valid, Silver is larger and still fully validated, while Bronze is the largest yet includes some non-valid pairs—allowing users to choose between size and reliability for downstream tasks.

\begin{table}[t]
\centering
\begin{tabular}{lrrr}
\toprule
Partition & 3 annot. & 2 annot. & 1 annot. \\
\midrule
Diamond & 195  & 356  & 0 \\ 
Gold    & 195  & 468  & 0 \\
Silver  & 195 & 468 & 2917 \\
Bronze  & 199 & 502 & 3296 \\
\bottomrule
\end{tabular}
\caption{Number of HS--CN pairs annotated by 3, 2 and 1 annotators.}
\label{tab:number-annotators-per-partiion}
\end{table}

Table \ref{tab:number-annotators-per-partiion} illustrates the distribution of annotated pairs based on the number of annotators involved. Within the Bronze partition, of the 199 pairs annotated by three annotators, only 4 pairs ($\approx$2.01\%) were deemed invalid by validators. For the 502 pairs annotated by two annotators, 34 pairs ($\approx$6.77\%) were marked invalid. Lastly, among the 3,296 pairs annotated by a single annotator, 379 pairs ($\approx$11.49\%) were classified as invalid. These observations indicate a clear trend: pairs annotated by multiple annotators tend to have proportionally fewer invalid instances, underscoring how reliability significantly improves with increased annotator agreement.
\todo{este parrafo es muy importante, intentar armar un plot o una tabla con este parrafo, para destacarlo.}
\todo{puedo agregar la conclusion del final de este parrafo a las conclsuiones finales de esta seccion}

\todo{Comento estos plot para ahorrar espacio. El de Gold se parece al de Diamond y el de Silver al de Bronze}

\todo{Puedo agregar un hatMap de cnType vs hsSubType, de todos los pares que usamos para anotacion -ya escribi un para hacer este heatMap mas arriba- y ver si tienen una distribucion similar a la de las cuatro particiones, eso explicaría lo que dice en "is curious..."}

\todo{Comento estos plot para ahorrar espacio. El de Gold se parece al de Diamond y el de Silver al de Bronze}

\todo{Must: comento este parrafo, porque florian me pidio info estadistica sobre esto. Abajo genere un parrafo con GPT describiendo las relaciones estadisticas de gender y education, pero como no estoy seguro de lo que dice, por ahora decido omitirlo (ademas de que esta seccion ya es suficientemente tediosa y de qu no estoy mostrando los plots para education ni gender):
Parrafo original:
We observed no significant differences between partitions based on gender or education; further analysis is omitted for brevity and limited interpretability.

Parrafo de GPT (creado en la charla "NII paper-> Hate Speech Counter-Narratives Dataset"):
Gender appears evenly balanced across partitions (χ² = 6.12, d.f.=3, p = 0.11; Cramér V = 0.03), indicating no detectable association between gender and partition membership.
By contrast, educational level shows a statistically reliable—though practically small—relationship with the partitions (χ² = 206.81, d.f.=6, p ≈ 6.7 × 10⁻⁴²; Cramér V = 0.11). Bachelor-level annotators make up just under half of each partition. Pairs contributed by annotators with Master's degrees are proportionally more frequent in Silver and Bronze (≈ 49 \%) than in Diamond and Gold (≈ 4 \%), whereas pairs from annotators with Some university without degree are more common in Diamond and Gold (≈ 10 \%) than in Silver and Bronze (≈ 1 \%). Although this skew is statistically significant, the small effect size suggests it is unlikely to bias downstream analyses appreciably.
}

In sum, the four-tier partitioning balances label reliability. Diamond and Gold deliver perfect validation, making them ideal for benchmarking model performance under minimal label noise. Silver adds scale without compromising valid pairs, while Bronze boosts volume, introducing the only subset of non-valid pairs. Altogether, the new annotations help fill clear gaps in the original CONAN dataset.

\todo{Really nice to have: estaria muy bueno analizar cuantos de los valid y cuántos de los non valid fueron anotados por 1, 2 o 3 anotadores, para poder dar una conclusioncomo esta:(i) multiple annotators markedly reduce invalid labels, }
\todo{REally nice to have: si hago el heatmap de todos los pares que fueron evaluados durante la anotacion, podria poner esta frase en el ultimo parrafo de conclusion: Bronze contributes the largest volume of HS–CN pairs and, despite containing the only non-valid subset, still retains informative patterns; non/ valid examples cluster in a handful of topic–strategy combinations, suggesting that annotation disagreements are systematic rather than random.}
\todo{Really nice to have: Intentar agregar una explicaciona  esta frase, me parece que una de las explicaciones es que la Silver y Diamond partition se parecen mas al conjunto de todos los pares que consideramos para anotar y que esta es la distribucion de todos los pares que consideramos para anotar. Para esto tendria que agregar la distribucion de educationLevel para todos los pares que consideramos para anotar al plot de la figura 15.}
\todo{Agregar algo como el Camer V para education level vs appropriateness.}

\todo{Puedo analizar cant de annotators vs validity y agregarlo al paper}

\todo{Si hago el analisis de appropriate vs education, puedo analizar esta conclusion para ver si la dejo como esta o la contradigo "This supports the intuition that crafting nuanced counter-speech benefits from higher formal education."}

\todo{Really nice to have: Esto lo puedo agregar cuando hable de las CN de tipo humor, pero corrigiendo las cantidades: whereas purely factual or denouncing replies fare markedly better. Interestingly, adding a mild humorous twist after an explicit condemnation (denouncing + humor) doubles the acceptance rate compared to humour alone."}

\todo{Asegurarme de que hago una comparación de heatmaps entre particioens: ej: appropriate bronze vs non-appropriate silver, non-valid bronze vs appropriate silver -notar que estoy comparando silver contra bronze-}
\todo{Must: antes esto estaba como lo comentado, pero como ahora tengo un solo experimento, no tiene sentido tener una "experimental evaluation" y una division en "evaluation of recommendation systems"}
\todo{Must: las secciones sec:ExperimentalEvaluation y sec:evaluationOfRecommendationSystems se unieron en la seccion sec:ExperimentalEvaluationEvaluationOfRecommendationSystems asi que tnego que arreglar todas las referencias a esto}

\subsection{Experimental Evaluation: Evaluation of Recommendation Systems.}
\label{sec:ExperimentalEvaluationEvaluationOfRecommendationSystems}

\todo{Must: agregar que pongo disponibles los datasets usados para este experimento. Estan en (con guiones bajos en vez de altos): My Drive niiAnnotationProjectAnnotatedExamples datasets-used-for-recommender-system-experiment. Tamaño de los datasets: 
conan-not-in-bronze-train-df: 3,119 rows
candidate-pool: 100 rows
bronze-hs: 45 rows
bronze-cn: 100 rows -> es lo mimso que candiate pool. 

Segun GPT esto hago con el dataset conan-not-in-bronze-train-df:
 - it contains all rows from the English-only CONAN dataset whose HS and CN are absent from the Bronze partition.
  -  Used to fit the TF-IDF vocabulary or to optionally fine-tune SBERT. -- This guarantees zero lexical overlap between anything seen during training and the evaluation ground truth.
}
\todo{Must: Esta nota (que me dijo GPT) puede ser importante para esta seccion (creo que la voy a tener que agregar): The only Bronze content used during training is the *vocabulary* of the candidate pool when SBERT encodes or BM25 tokenises those CNs, which is fine because the labels remain hidden.

Y dps me dijo: Training/representation learning uses only `conan_not_in_bronze_train_df`

Esto hace referecia a este snippet del experimento donde armo los recommenders (ojo que le saque algunos guiones bajos):
systems={}
for name,builder in builders.items():
    systems[name]=builder(conan not in bronze train df, candidate pool)
}
\todo{Must: en esta seccion voy a tener que aclara que para estsos experimentos asumo que si un para no esta etiquetado como appropriate, entonces asumo que es non-appropriate (puede suceder que no este como appropriate porque en realidad esta anotado como '1: no lo se', pero para este experimento asumoque si no esta como appropriate -> esta como non-appropriate)}
\todo{Must: en todo el paper tengo que unificar si escribo \texttt{cnType}  o solo cnType y lo mismo para hsSubType.}
\todo{Must: Algunos de los que tengo que hacer si o si, los marco con "must".}

\todo{Must: Esto tambien lo agrego ahora, ver si voy a escribir hsSubType o poner HS sub Type, lo mimso para cnType y CN type}

As discussed in §\ref{sec:Introduction}, incomplete CONAN labels masked appropriate CNs. In what follows, we re-evaluate recommendation systems on our exhaustively annotated FC-CONAN partitions\footnote{The datasets used in this experiment are available at \url{https://github.com/jnqeras/FC-CONAN-dataset/tree/main/recommender_experiment_data}}.

We compare twelve recommenders trained on an English-only dataset (\textit{conan\_not\_in\_bronze\_train}) created by excluding any HS or CN found in the Bronze partition, ensuring no overlap between training and evaluation. Specifically, we evaluate:
\setlength\itemsep{0pt}        
\begin{itemize}
  \item \textbf{TF--IDF} – cosine similarity on TF--IDF vectors (sparse baseline); \citep{SALTON1988513}.
  \item \textbf{BM25} – Okapi BM25 lexical ranker ($k_{1}{=}1.2$, $b{=}0.75$); \citep{RobertsonStephen}.
  \item \textbf{random} – uniform sampling of ten candidate CNs.
  \item \textbf{sbert(MiniLM)} – Sentence-BERT \textit{all-MiniLM-L6-v2} (384 d) + cosine; \citep{sBertGurevych}, \citep{10.5555/3495724.3496209}.
  \item \textbf{sbert(MPNET)} – Sentence-BERT \textit{all-mpnet-base-v2} (768 d) + cosine; \citep{10.5555/3495724.3497138}.
  \item \textbf{sbert(RoBERTa-L)} – Sentence-BERT \textit{all-roberta-large-v1} (1024 d) + cosine; \citep{zhuang-etal-2021-robustly}.
  \item \textbf{textEmb3L} – OpenAI dense encoder \textit{text-embedding-3-large} (3 072 d); \citep{openai2024newembedding}.
  \item \textbf{GPT-4o} – zero-shot prompting of \textit{gpt-4o-2024-08-06} to return ten CNs; \citep{openai24gpt4o}.
  \item \textbf{RR25 system 1 / RR50 system 1} – top-25/50 CNs from textEmb3L, then GPT-4o re-ranks.
  \item \textbf{RR25 system 2 / RR50 system 2} – top-25/50 CNs from sbert(MPNET), then GPT-4o re-ranks.
\end{itemize}
To ensure a fair comparison, each system ranks a fixed candidate pool: the 100 unique CNs from the Bronze partition. This simulates a deployment where all models share the same, unchanging reply catalogue. \todo{Confirmar que la bronze partition tiene 100 CN distintas.} To maintain consistency, we limit our query set to the hate speech instances in the Bronze partition (\textit{bronze\_HS}). For each HS in \textit{bronze\_HS}, each recommender produces its top-10 counter-narratives, yielding one ranking per model.

We then evaluate each ranking against all four partitions (Diamond, Gold, Silver, and Bronze). A recommendation is counted as correct if —and only if— the \(\langle \text{HS};\text{CN}\rangle\) pair is annotated as appropriate in the partition under test. As an example, we show the TF-IDF ranking for the following HS (capped at 2 CNs to keep the paper concise):

\begin{itemize}
    \item "When Islam hasn't arrived yet in England, things was good. Now we are full of muslim rapists and it is evident that sexual violence is jihad in England".
\end{itemize}

The ranking is marked \faCheck \space or \faTimes \space for each partition to illustrate how appropriateness varies depending on which annotation standard we use.
\todo{unificar si en el paper las llamo partitions o standard dataset}
\todo{Unificar el nombre del modelo a TDIDF o a como este en los plots.}
    
\begin{enumerate}
    \item Muslims aren't rapists or murderers, in fact many Muslims are doing good things to fight the extremists falsely appropriating Islam: https://5pillarsuk.com/2019/01/03/muslim-woman-praised-for-aiding-victims-of-new-years-eve-terror-attack-in-manchester/.   [D\faTimes, G\faTimes, S\faTimes, B\faTimes]
    \item Rapists are, in most cases, friends, family, or partners of the victim. The proportion of Pakistani people who are rapists is no more than that of White British population.   [D\faTimes, G\faCheck, S\faCheck, B\faCheck]
    \todo{Must: estos elementos del ranking estan comentados para hacer el ranking mas corto y el paper sea mas corto. Acomodar las referencias a este ranking para que digan que tienen 5 en vez de 10 contranarrativas.}
\end{enumerate}

\todo{estos hs y cn estan en todas las d, g, s and b partitions y si tengo tiempo chequear que los pares que tienen un tick, estan en cada una de las DGSB partitions.}

In the example ranking, the top-ranked counter-narrative (CN) is never judged appropriate in any partition. By contrast, the second-ranked CN is judged appropriate in the Gold, Silver, and Bronze partitions, but not in Diamond. This demonstrates that the evaluation metrics computed on a generated ranking can change substantially depending on which partition is used to define “appropriate” pairs.

\begin{figure}[!t]
\centering
\includegraphics[width=1\columnwidth]{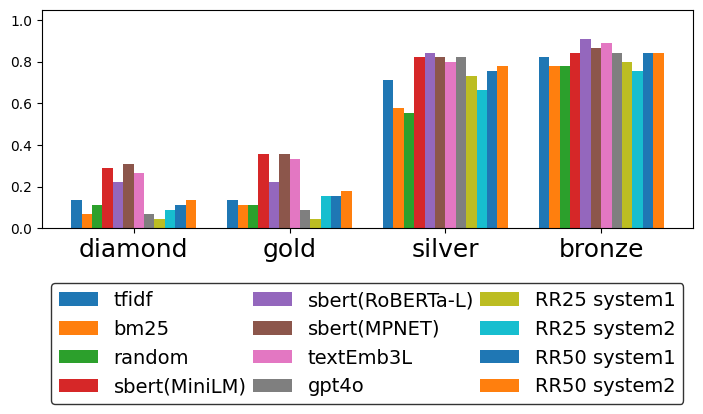}
\caption{HIT RATIO@10 across the twelve systems (glossary in Section \ref{sec:ExperimentalEvaluationEvaluationOfRecommendationSystems}).}
\label{fig.16}
\end{figure}

\todo{Must: Esta imagen queda muy chica, voy a tener que editarla de alguna forma. Sacar el titular de adentro de la imagen y ponerla en el caption, las notas e los ejes (count y cnType), las cnType abajo de cada barra y los nombres de las particiones, estan muy chicas. Preguntar a GPT como mostrar esto mas grande. Este fixme aplica a todos los plots de esta seccion (y quizas a algunos otros mas).}

\begin{figure}[!t]
\centering
\includegraphics[width=1\columnwidth]{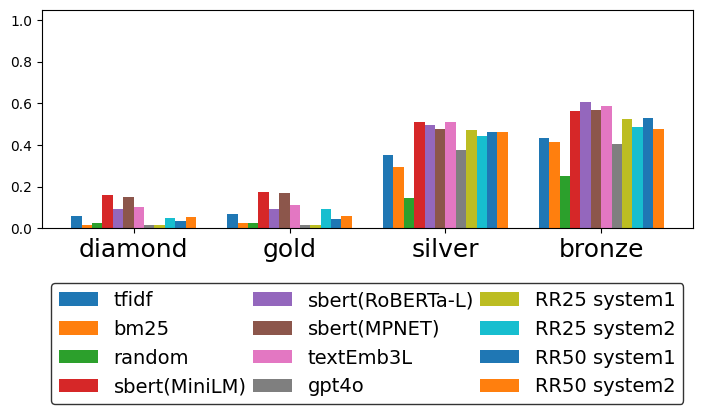}
\caption{MRR@10 of the twelve evaluated systems (see glossary in Section \ref{sec:ExperimentalEvaluationEvaluationOfRecommendationSystems}).}
\label{fig.20}
\end{figure}
\todo{arreglarle el caption a las imagenes.}
\todo{Precision at k y accuracy at k tienen literalmente el mismo plot, tengo que revisar en la notebook en donde los genero porque me parece que estoy generando literalmente el mismo plot. Cuando arregle esto, tengo que arreglar lo que digo sobre estas metricas.}

Figures \ref{fig.16} and \ref{fig.20} show that recommender systems performance improves with partitions containing a greater number of annotated “appropriate” pairs: Diamond scores lowest, followed by Gold, then Silver, with Bronze achieving the highest values across both metrics. Overall, metric values scale roughly in proportion to the number of appropriate pairs in each partition. We observe the same trend for metrics such as NDCG@10, MAP@10, Precision@10, Accuracy@10, and F1@10, although their plots are omitted due to space limitations.

None of the pairs formed from the candidate pool (Bronze CNs) and \textit{bronze\_HS} are labeled as appropriate in the CONAN dataset; thus, using CONAN as the sole gold standard for this set yields zero scores across all metrics. As progressively more appropriate pairs are included—from CONAN’s subset to Diamond, Gold, Silver, and Bronze—metric scores increase. This suggests that when appropriate pairs remain unannotated, evaluation metrics serve only as lower bounds and fail to reflect true retrieval system performance.

\begin{table}[t]
\centering
\begin{tabular}{lrrrr}
\toprule
System  & Avg. & Min. & Max. & CV\% \\
\midrule
sbert(MPNET) & 0.3283 & 0.1836 & 0.4904 & 38.425     \\
sbert(MiniLM) & 0.3270 & 0.1766	& 0.4804 & 39.200     \\
textEmb3L & 0.3128 & 0.1366	& 0.5098 & 53.200     \\
sbert(RoBERTa-L) & 0.3034 & 0.1214 & 0.5138	& 56.400     \\
RR50 system2 & 0.2441 & 0.0658 & 0.4271 & 69.175     \\
RR50 system1 & 0.2416 & 0.0526 & 0.4508 & 75.350     \\
RR25 system2 & 0.2278 & 0.0514 & 0.3992 & 63.175     \\
TF-IDF & 0.2203 & 0.0703 & 0.4024 & 63.275     \\
RR25 system1 & 0.2165 & 0.0184 & 0.4339 & 91.950     \\
GPT-4o & 0.2154 & 0.0267 & 0.4118 & 87.100     \\
BM25 & 0.1819 & 0.0296 & 0.3810 & 81.650     \\
random & 0.1500 & 0.0414 & 0.3115 & 78.100     \\
\bottomrule
\end{tabular}
\caption{
Macro-level robustness of the 12 systems.  
For each system we average, over four metrics (HIT RATIO@10, MRR@10, NDCG@10, MAP@10), the metric–wise \textit{mean}, \textit{minimum}, \textit{maximum} and coefficient of variation (CV\% = $(\text{standard deviation}/{\text{mean}}) \times 100$). Higher “Avg.” indicates better overall effectiveness, while lower CV\% indicates greater stability across the Diamond–Bronze partitions.}
\label{tab:recommendersRobustness}
\end{table}

Table \ref{tab:recommendersRobustness} shows a clear performance hierarchy. 
\textbf{Embedding-based rankers} (all SBERT variants plus OpenAI’s \texttt{textEmb3L}) obtain the highest average score ($\mu\!\approx\!0.32$) and the lowest coefficient of variation ($CV\%\!\approx\!46$), indicating that dense vector representations are both \textit{effective} and \textit{robust} to missing annotations. 
\textbf{Hybrid rerankers} (RR50 system 1/2 and RR25 system 2) come next in terms of performance ($\mu\!\approx\!0.23$) yet remain substantially less stable ($CV\%\!\approx\!69$), presumably because the GPT-4o reranking step amplifies noise whenever the embedding pre-filter retrieves weak candidates. 
Among the \textbf{lexical baselines}, TF–IDF matches hybrid effectiveness ($\mu\!\approx\!0.22$) while BM25 lags behind ($\mu\!\approx\!0.18$); both exhibit high variability ($CV\%\!>\!63$), confirming their sensitivity to annotation sparsity. 
The \textbf{LLM zero-shot} strategy (GPT-4o alone) clusters with the Hybrid reranker RR25 system 1 ($\mu\!\approx\!0.21$, $CV\%\!\approx\!89$). 
As expected, the \textbf{random} baseline sits at the bottom ($\mu\!=\!0.15$). 
Overall, the results support the conclusion that \emph{representation quality—rather than pipeline complexity alone—drives both effectiveness and robustness} in counter-narrative retrieval.

\todo{Creo que estoy usando los nombres resumidos de los modelos, por ejemplo MPNet deberia ser Sbert-MPNet}
\todo{aca no menciono a f2@k agregar si hace falta.}
\todo{Chequear si la meterica que tengo es "macro-averaged Precision@10" o solamente "Precision@k", si es el primero, tengo que cambiar el titulo a la figura de esta metrica.}
\todo{Chequear que MRR es MRR@10 como dice la imagen y el texto}

\todo{Creo que falta agregar un parrafo de conclusion diciendo que nuestra hipotesis se cumple. Y si no, agregar esta conclusion a la seccion conclusiones -haga lo que haga, tengo que hacer lo mismo para el experimento generativo-. Esta puede ser una conclusion: "Partitions with more annotated 'appropriate' pairs yield higher metric scores. In other words, if the corpus is under-annotated, the evaluation metrics will underestimate the system's true performance." lo que viene antes del "in other words" ya lo escribi en esta seccion.}

\todo{Buscar paper de Virginia Villata -creo que era de ella- en e cual dice que 0.6 para metricas de sistemas de recomendacion para discrusos de odio estaban muy bien, para ver si alguna particion llega a ese valor.}
\todo{Ver que el texto de esta sección se genere apropiadamente.}
\todo{Confirmar si los modelos son rerieval-based, como dice el texto.}
\todo{Confirmar si los nombres de los modelos están bien}
\todo{Agregar citas a cada sistema de recomendacion.}

\todo{Must: esta seccion de text generation la comento porque no entra para la version para el workshop.}
\nocite{*} \todo{quizas sacar esto}

\section{Limitations and Future Work}
\label{sec:limitations}
\todo{Must: en algun lugar de esta seccion de Future Work, tengo que agregar que tengo que hacer experimentos de contrastive learning para aprovechar el full potential de este dataset -ver si se puede hacer finetunning de LLM con contrastive learning-. }

\textbf{Generative fine-tuning and contrastive learning:} Generative models can be fine-tuned on appropriate pairs from each partition to compare performance—we conducted such experiments but omitted them due to space limitations—while contrastive learning could leverage both appropriate and non-appropriate pairs to fully exploit the dataset’s structure.

\textbf{Language scope:} We cover only the English portion of CONAN. Extending exhaustive pairing to French and Italian remains future work.

\textbf{Annotation coverage:} Our 45 HS × 129 CN subset produced partitions large enough to reveal evaluation artefacts, yet remains far from a \emph{fully} exhaustive re-annotation of all possible combinations of CONAN. A semi-automatic “LLM-first, human-verify” pipeline could finish that job at lower cost.

\todo{must: comento este item porque refiere a los experimentos generativos (que no estan incluidos en el paper)}

\textbf{Demographic diversity:} Our annotator pool is skewed towards young, highly educated English-speaking individuals. Broader demographic sampling would reveal whether cultural background influences appropriateness judgments.

\section{Conclusions}
\label{sec:conclusion-final}
\todo{Must: el CONAN tiene 4078 pares contando los tres idiomas (chequear), ver cuantos tiene en ingles y compararlo contra mi dataset.}
We introduced \textbf{FC-CONAN}, to the best of our knowledge, the \emph{first} hate speech / counter-narrative dataset where \emph{every} possible pairing between two finite sets—45 HS messages and 129 CNs—is explicitly judged.  \todo{si le cambio el nombre al dataset, cambiarlo de aca.} \todo{Unificar como llamo a los hate speech y counter-narrative} \todo{confirmar que esa es la cantidad de HS y CN que combinamos para crear el datset.}

Consequently, if a pair is labeled \textit{appropriate} in a given partition, it means that—according to the requirements of that partition—it was indeed deemed suitable. Conversely, if it is labeled \textit{non-appropriate}, it reflects that it did not meet those same partition-specific criteria.

Four quality-controlled partitions—\textsc{Diamond}, \textsc{Gold}, \textsc{Silver}, and \textsc{Bronze}—let practitioners trade annotation reliability for corpus size.

None of the HS–CN pairs we annotated occurs in the original CONAN corpus; our partitions therefore add hundreds of previously missing \emph{appropriate} CNs, revealing how many unlabeled appropriate pairs the sparse labels in CONAN contained. 

\textbf{Importantly}, using the partitions introduced, we confirm that not considering all possible pairs during dataset creation leaves many appropriate pairs unannotated. This results in artificially lower scores for most evaluation metrics in recommendation systems. For metrics that are not negatively affected by a greater number of positives—such as MRR@10 or Hit Ratio@10—this implies that scores obtained when evaluating on partially annotated datasets should be considered \emph{lower bounds}.

We also showed that embedding-based rankers outperform others in both effectiveness and robustness to missing annotations.

In summary, researchers evaluating counter-narrative \emph{retrieval} systems should rely on densely annotated datasets such as ours to avoid underestimating system performance.

\todo{Agregar como future work que la data puede ser usada para entrenar llms para anotar dataset as grandes. Otra idea es usar un LLM para terminar de anotar todas las combinaciones del CONAN dataset.}

\section*{Acknowledgements}

This work was supported by the “R\&D Hub Aimed at Ensuring Transparency and Reliability of Generative AI Models” project of the Ministry of Education, Culture, Sports, Science and Technology and JSPS KAKENHI Grant Numbers, JP22H00543. We also sincerely thank the annotators and validators who generously volunteered their time to contribute to this project. Junqueras was partially supported by the UBA BIICC Fellowship Program, the Fundar FunDatos Fellowship Program, and the NII International Internship Program.

\todo{Must: viejo ack: We thank the nine annotators and four validators for their diligence.} \todo{Must: Agregar esto solamente si el grupo me confirma que lo puedo agregar
... and Satoh sensei for valuable feedback on disagreement metrics
This work was supported by NII internship grant.}

\bibliographystyle{kr}
\bibliography{kr-sample}

\end{document}